\documentclass[journal]{IEEEtai}
\usepackage{amsmath,amsfonts,bm}

\def\eqref#1{equation~\ref{#1}}

\def\1{\bm{1}}

\def\rvepsilon{{\bm{\epsilon}}}

\def\mI{{\bm{I}}}

\def\mSigma{{\bm{\Sigma}}}

\DeclareMathAlphabet{\mathsfit}{\encodingdefault}{\sfdefault}{m}{sl}
\SetMathAlphabet{\mathsfit}{bold}{\encodingdefault}{\sfdefault}{bx}{n}

\def\gN{{\mathcal{N}}}

\newcommand{\E}{\mathbb{E}}

\DeclareMathOperator*{\argmin}{arg\,min}

\newcommand{\sbr}[1]{\left[#1\right]}
\newcommand{\given}{\,|\,}

\usepackage{hyperref}
\usepackage{url}

\usepackage{tabularray} 
\UseTblrLibrary{booktabs}
\usepackage{color, colortbl}
\usepackage{mathtools}
\usepackage{cleveref}
\usepackage{bbm}
\usepackage[ruled,linesnumbered]{algorithm2e}
\usepackage{amsmath}
\usepackage{subcaption}
\usepackage{multirow}
\usepackage{amssymb}
\usepackage[font=small]{caption}
\usepackage{hyperref} 
\usepackage{cite}
\usepackage{color}

\title{\LARGE \bf
DDM-Lag : A Diffusion-based Decision-making Model for Autonomous Vehicles with Lagrangian Safety Enhancement
}

\author{Jiaqi Liu,~\IEEEmembership{Student Member,~IEEE,} Peng Hang,~\IEEEmembership{Member,~IEEE,} Xiaocong Zhao, Jianqiang Wang, and Jian Sun % <-this % stops a space
\thanks{This work was supported in part by the National Natural Science Foundation of China (52125208, 52232015), the Belt and Road Cooperation Program under the 2023 Shanghai Action Plan for Science, Technology and Innovation (23210750500), the Shanghai Scientific Innovation Foundation (23DZ1203400) and the Fundamental Research Funds for the Central Universities.}% <-this % stops a space
\thanks{Jiaqi Liu, Peng Hang, Xiaocong Zhao, and Jian Sun are with the Department of
Traffic Engineering and Key Laboratory of Road and Traffic Engineering,
Ministry of Education, Tongji University, Shanghai 201804, China. (e-mail: \{liujiaqi13, hangpeng, zhaoxc, sunjian\}@tongji.edu.cn)}
\thanks{Jianqiang Wang is with State Key Laboratory of Automotive Safety and Energy
School of Vehicle and Mobility, Tsinghua University, Beijing
100084, China.(e-mail: wjqlws@tsinghua.edu.cn)}
\thanks{Corresponding author: Jian Sun}
}

\begin{document}

\maketitle
\thispagestyle{empty}
\pagestyle{empty}

%%%%%%%%%%%%%%%%%%%%%%%%%%%%%%%%%%%%%%%%%%%%%%%%%%%%%%%%%%%%%%%%%%%%%%%%%%%%%%%%
\begin{abstract}
Decision-making stands as a pivotal component in the realm of autonomous vehicles (AVs), playing a crucial role in navigating the intricacies of autonomous driving. Amidst the evolving landscape of data-driven methodologies, enhancing decision-making performance in complex scenarios has emerged as a prominent research focus.
Despite considerable advancements, current learning-based decision-making approaches exhibit potential for refinement, particularly in aspects of policy articulation and safety assurance. 
To address these challenges, we introduce DDM-Lag, a Diffusion Decision Model,  augmented with Lagrangian-based safety enhancements.
This work conceptualizes the sequential decision-making challenge inherent in autonomous driving as a problem of generative modeling, adopting diffusion models as the medium for assimilating patterns of decision-making. We introduce a hybrid policy update strategy for diffusion models, amalgamating the principles of behavior cloning and Q-learning, alongside the formulation of an Actor-Critic architecture for the facilitation of updates. To augment the model's exploration process with a layer of safety, we incorporate additional safety constraints, employing a sophisticated policy optimization technique predicated on Lagrangian relaxation to refine the policy learning endeavor comprehensively. Empirical evaluation of our proposed decision-making methodology was conducted across a spectrum of driving tasks, distinguished by their varying degrees of complexity and environmental contexts. The comparative analysis with established baseline methodologies elucidates our model's superior performance, particularly in dimensions of safety and holistic efficacy.
\end{abstract}

\begin{IEEEImpStatement}
In an era where autonomous vehicles (AVs) symbolize the cutting edge of transportation innovation, ensuring the safety and reliability of their decision-making systems remains a pivotal challenge. Existing artificial intelligence methods, such as reinforcement learning, still lack sufficient progress in their large-scale safe application on autonomous vehicles, with key issues including the safety of data-driven approaches and the adaptability of their strategies. Our research introduces DDM-Lag, an advanced diffusion-based decision-making model with integrated Lagrangian safety enhancements, specifically designed for AVs. This method not only elevates the state of AV decision-making through the use of generative modeling and sophisticated optimization techniques but also significantly enhances safety and adaptability in dynamic environments. DDM-Lag contributes to elevating the intelligence level of decision-making in autonomous vehicles and provides a blueprint for applying similar methodologies in other domains requiring reliable decision-making under uncertainty. This study underscores the potential of merging advanced computational models with safety-centric optimizations to enhance the safe operation of intelligent systems.
\end{IEEEImpStatement}

%%%%%%%%%%%%%%%%%%%%%%%%%%%%%%%%%%%%%%%%%%%%%%%%%%%%%%%%%%%%%%%%%%%%%%%%%%%%%%%%
\section{Introduction}
The advent of autonomous driving technology is poised to transform the global transportation system fundamentally\cite{wang2023new,hang2022conflict}. Decision-making, a pivotal element within autonomous driving systems, plays an essential role in ensuring the safety, stability, and efficiency of autonomous vehicles (AVs)\cite{liu2023towards}. Despite significant advancements, the decision-making capabilities of AVs in complex scenarios are yet to reach their full potential. Concurrently, the evolution of deep learning has propelled learning-based methodologies to the forefront, with reinforcement learning (RL) being a particularly notable method. RL approaches sequential decision-making by optimizing the cumulative rewards of a trained agent. This method has demonstrated superior performance over human capabilities in various complex decision scenarios\cite{he2023robotic}. However, RL still faces challenges in decision safety, sampling efficiency, policy articulation, and training stability\cite{kiran2021deep,jeddi2023memory,ilahi2021challenges}.

In a different vein, diffusion models, a class of advanced generative models, have recently achieved remarkable success, especially in image generation\cite{croitoru2023diffusion}. Functioning as probabilistic models, they incorporate noise into data in a forward process and then iteratively remove the noise to recover the original data, following a Markov chain framework. Compared to other generative models, diffusion models excel in sampling efficiency, data fidelity, training consistency, and controllability\cite{croitoru2023diffusion,yang2023diffusion}. Moreover, these models have been validated as effective tools for enhancing decision-making in reinforcement learning strategies\cite{wang2022diffusion}. While diffusion models have seen applications in game strategy and robotic motion planning\cite{lu2023synthetic, he2023diffusion}, their potential in autonomous driving decision-making is still largely untapped. This exploration is imperative, especially considering the black-box nature of neural networks, which complicates the direct assurance of decision-making safety—a critical requirement for highly safety-conscious autonomous vehicles.

To address these challenges, we introduce DDM-Lag, a novel Diffusion Decision Model augmented with Lagrangian-based safety enhancements, specifically tailored for improving decision-making in autonomous driving. 

\begin{figure}
    \centering
    \includegraphics[width=1\linewidth]{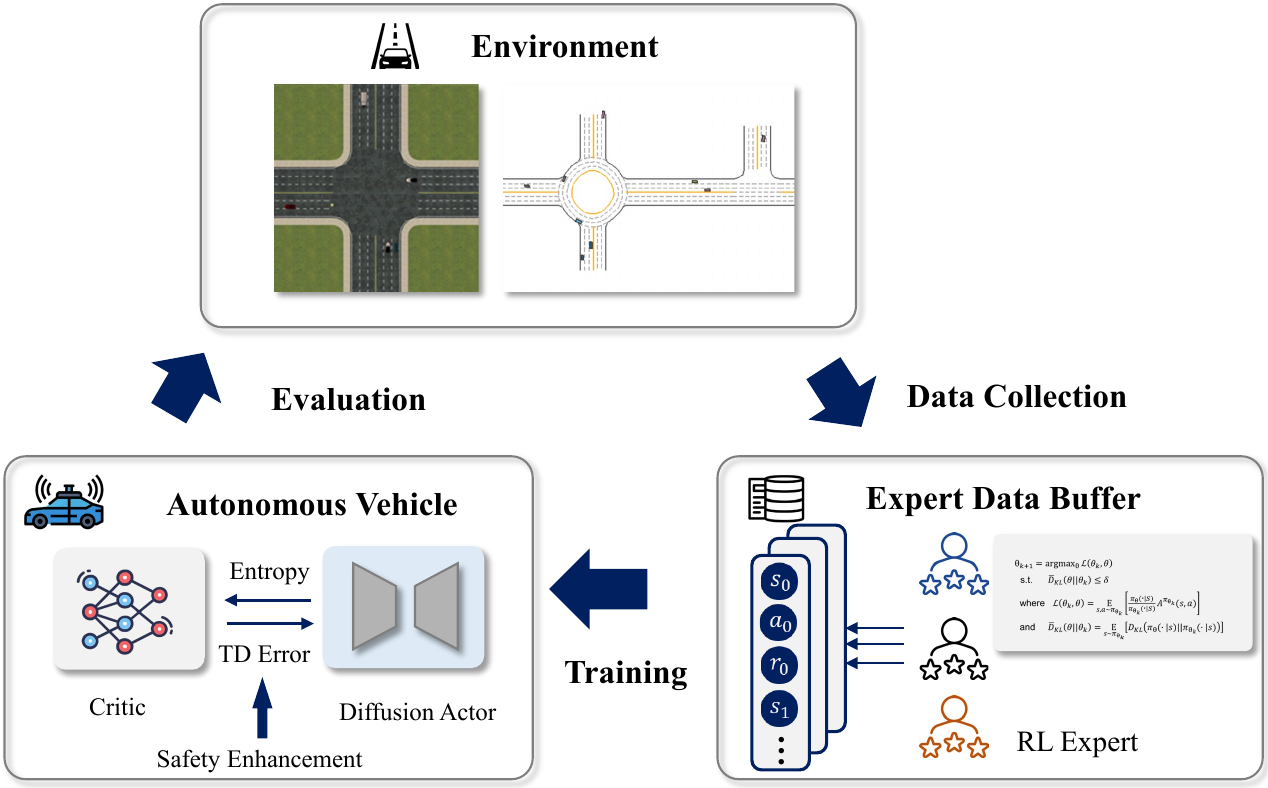}
    \caption{The overall procedure of our work.}
    \label{fig:overall_procedure}
\end{figure}

We commence by delineating the sequential decision-making quandary in autonomous driving as a Constrained Markov Decision Process (CMDP), adopting a perspective rooted in generative modeling. We integrate diffusion models for resolving the generative decision-making conundrum, where models ingest environmental perception information as input and output vehicle control variables, learning and updating through a generative autoregressive methodology. For the stable and efficient updating of the diffusion model, we propose a hybrid policy update method that amalgamates behavior cloning and Q-learning within a diffusion model framework, and correspondingly design an Actor-Critic architecture to facilitate this updating process. Within this update framework, our diffusion model's updating objectives include two components: 1) a behavior cloning term, encouraging the diffusion model to sample behaviors from a distribution analogous to the training set; and 2) a policy improvement term, aimed at sampling higher-value actions. Concurrently, additional safety constraints are integrated within the model's exploration constraints to ensure the safety of action exploration, employing a policy optimization methodology predicated on Lagrangian relaxation to refine the entire policy learning trajectory comprehensively.
The model update phase incorporates a Proportional-Integral-Derivative (PID) controller for the adjustment of $\lambda$, ensuring a stable update trajectory. Fig.\ref{fig:overall_procedure} delineates the entire workflow of our initiative, including the development of an offline expert data collection module. This module is pivotal for training multiple reinforcement learning agents and amassing data from diverse scenarios.

Finally, the efficacy of the DDM-Lag approach undergoes evaluation in various driving tasks, each differing in complexity and environmental context. When juxtaposed with established baseline methods, our model exhibits superior performance, particularly in aspects of decision-making safety and comprehensive performance.

Our contributions can be summarized as follows:
\begin{itemize}    
    \item We model the autonomous driving decision-making process as a generative diffusion process, proposing a hybrid Policy update method that integrates behavior cloning with Q-learning, facilitated by an Actor-Critic framework for policy updates.
    \item A Lagrangian relaxation-based policy optimization approach is adopted, enhancing the safety  of the decision-making process.
    \item The proposed method is subjected to  testing in a variety of driving scenarios, demonstrating advantages in safety and comprehensive performance.
\end{itemize}

The rest of the paper is organized as follows. Section~\ref{section:2} summarizes the recent related works. The preliminaries and the problem formulation are described in section~\ref{section:3}. In section~\ref{section:4}, the framework we proposed is described. The simulation environment and comprehensive experiments are introduced and the results are analyzed in section \ref{section:5}. Finally, this paper is concluded in section \ref{section:6}.

\section{Related Works}
\label{section:2}
\subsection{Decision-Making of Autonomous Vehicles}

The decision-making process is a cornerstone in the functionality of autonomous vehicles (AVs). Traditional approaches in this realm encompass rule-based, game theory-based, and learning-based methodologies.
Significantly, with the advent and rapid progression of deep learning and artificial intelligence, learning-based approaches have garnered increasing interest. These methods, particularly reinforcement learning (RL) \cite{saxena2020driving,li2020deep} and large language models (LLM) \cite{wen2023dilu,chen2023driving}, exhibit formidable and extraordinary learning capacities. They adeptly handle intricate and dynamic environments where conventional rule-based decision-making systems may falter or respond sluggishly \cite{schwarting2018planning}.

Saxena et al. \cite{saxena2020driving} introduced a pioneering model-free RL strategy, enabling the derivation of a continuous control policy across the AVs' action spectrum, thus significantly enhancing safety in dense traffic situations. Liu et al. \cite{liu2023mtd} developed a transformer-based model addressing the multi-task decision-making challenges at unregulated intersections. Concurrently, LLMs have also emerged as a focal point in AV decision-making. The Dilu framework \cite{wen2023dilu}, integrating a Reasoning and Reflection module, facilitates decision-making grounded in common-sense knowledge and allows for continuous system evolution.

\subsection{Reinforcement Learning in Safe Decision-Making}

In contemporary research, reinforcement learning is recognized as an efficacious learning-based method for sequential decision-making\cite{liu2024enhancing}. By optimizing the cumulative reward, RL identifies the optimal action strategy for agents \cite{he2023robotic}, finding applications in domains such as robotic control, gaming, and autonomous vehicles \cite{wang2020deep}. However, the 'black box' nature of RL makes the safety of its policy outputs challenging to guarantee\cite{jeddi2023memory}.

To augment the safety aspect in RL decision-making, various methodologies have been proposed to incorporate safety layers or regulate the agents' exploratory processes during training \cite{gu2022review,jeddi2023memory}. Safe RL is often conceptualized as a Constrained Markov Decision Process (CMDP) \cite{altman2021constrained}, incorporating constraints to mitigate unsafe exploration by the agent. Borkar \cite{borkar2005actor} proposed an actor-critic RL approach for CMDP, employing the envelope theorem from mathematical economics and analyzing primal-dual optimization through a three-time scale process. Berkenkamp et al. \cite{berkenkamp2017safe} developed a safe model-based RL algorithm using Lyapunov functions, ensuring stability under the assumption of a Gaussian process prior.

\subsection{Applications of the Diffusion Model}

The diffusion model has emerged as a potent generative deep-learning tool, utilizing a denoising framework for data generation, with notable success in image generation and data synthesis \cite{croitoru2023diffusion}. Recent studies have applied the diffusion model to sequential decision-making challenges, functioning as a planner \cite{janner2022planning,ajay2022conditional}, policy network \cite{wang2022diffusion,he2023diffusion}, and data synthesizer \cite{lu2023synthetic, he2023diffusion}.

Diffuser \cite{janner2022planning} employs the diffusion model for trajectory generation, leveraging offline dataset learning and guided sampling for future trajectory planning. Wang et al. \cite{wang2022diffusion} demonstrated the superior performance of a diffusion-model-based policy over traditional Gaussian policies in Q-learning, particularly for offline reinforcement learning. Additionally, to enhance dataset robustness, Lu et al. \cite{lu2023synthetic} utilized the diffusion model for data synthesis, learning from both offline and online datasets.

In our study, we posit the diffusion model as a central actor in the decision-making process of autonomous agents, aiming to augment the flexibility and diversity in AV decision-making strategies.

\section{Preliminaries}
\label{section:3}
\subsection{Constrained Markov Decision Process (CMDP)}
A CMDP is characterized by the tuple $(\mathcal{S},\mathcal{A},\mathcal{R},C,\gamma,\mu)$, where $\mathcal{S}$ represents the state space, $\mathcal{A}$ denotes the action space, $\mathcal{R}: \mathcal{S} \times \mathcal{A} \rightarrow \mathbb{R}$ is the reward function, and $C: \mathcal{S} \times \mathcal{A}  \rightarrow \mathbb{R}$ signifies the corresponding single-stage cost function. The discount factor is denoted by $\gamma$, and $\mu$ indicates the initial state distribution.
We define a policy $\pi$ as a map to from states to a probability distribution over actions, and $\pi(a|s)$ represents the probability of action $a$ based on state $s$.

% We define a policy $\pi = \{\pi_0, \pi_1, \ldots\}$ as a sequence of decision rules for action selection. Specifically, for any $k \geq 0$ and $s \in S$, $\pi_k(s) \in \mathbb{P}(s)$ represents the probability distribution $\pi_k(s) \stackrel{\triangle}{=} (\pi_k(s,a), a \in A(s))$, with $\pi_k(s,a)$ being the likelihood of choosing action $a$ in state $s$ at time $k$ under policy $\pi$. Here, $A(s)$ is the set of viable actions in state $s$, hence ${\displaystyle A = \cup_{s \in S} A(s)}$. Such a policy is often termed a randomized policy. In contrast, a stationary policy is a randomized policy where $\pi_k = \pi_l$ for all $k \neq l$, meaning it selects actions based on a fixed distribution irrespective of the time of action selection. We denote a stationary policy also as $\pi$ for simplicity.

Our study focuses on a class of stationary policies $\pi_\theta$, parameterized by $\theta$. The objective function is framed in terms of the infinite horizon discounted reward criterion: $\mathbb{E}\left[\sum_{t=0}^{\infty} \gamma^t R(s_t,a_t) \mid s_0 \sim \mu, a_t \sim \pi_\theta, \forall t\right]$. Similarly, the constraint function is expressed via the infinite horizon discounted cost: $\mathbb{E} \left[\sum_{t=0}^{\infty} \gamma^t C(s_t,a_t)\mid s_0 \sim \mu, a_t \sim \pi_\theta, \forall t\right]$. The optimization problem with constraints can be formulated as follows:
\begin{equation}
\begin{aligned}
    \label{cop1}
    \max_\theta \mathbb{E}\left[\sum_{t=0}^{\infty}
    \gamma^t R(s_t,a_t) \mid s_0\sim \mu, a_t \sim \pi_\theta, \forall t\right]. \\ 
    \ \mbox{s.t.}\  \mathbb{E} \left[\sum_{t=0}^{\infty}
    \gamma^t C(s_t,a_t)\mid s_0\sim \mu, a_t \sim \pi_\theta, \forall t\right] \leq d.
\end{aligned}
\end{equation}

\subsection{Lagrangian Methods}
In the context of the constrained optimization problem as delineated in \eqref{cop1}, the associated Lagrangian can be articulated as:
 \begin{equation}
 \label{lagrangian}
        L(\theta, \lambda) = J^R(\pi_\theta) - \lambda (J^C(\pi_\theta) - d)
 \end{equation}
where $\lambda \in \mathbb{R}^+$ denotes the Lagrange multiplier, $J^R(\pi_\theta)= \mathbb{E}\left[\sum_{t=0}^{\infty} \gamma^t R(s_t,a_t,s_{t+1}) \mid s_0 \sim \mu, a_t \sim \pi_\theta, \forall t\right]$, $J^C(\pi_\theta) = \mathbb{E} \left[\sum_{t=0}^{\infty} \gamma^t C(s_t,a_t,s_{t+1})\mid s_0 \sim \mu, a_t \sim \pi_\theta, \forall t\right]$. The objective is to identify a tuple $(\theta^*,\lambda^*)$ that represents both the policy and the Lagrange parameter, fulfilling the condition:
\begin{equation}
    \label{lagrangeoptim}
     L(\theta^*,\lambda^*) = \max_\theta \min_\lambda L(\theta,\lambda).
\end{equation}
Resolving the max-min problem is tantamount to locating a global optimal saddle point $(\theta^*,\lambda^*)$ such that for all $(\theta,\lambda)$, the following inequality is maintained:
\begin{equation}
    \label{saddlepointcondition}
 L(\theta^*, \lambda) \geq L(\theta^*, \lambda^*) \geq  L(\theta, \lambda^*).
\end{equation}
Given that $\theta$ is associated with the parameters of a Deep Neural Network, identifying such a globally optimal saddle point is computationally challenging. Thus, our objective shifts to finding a locally optimal saddle point, satisfying \eqref{saddlepointcondition} within a defined local neighbourhood $H_{\epsilon_1,\epsilon_2}$:
\begin{equation}
\label{H}
          H_{\epsilon_1,\epsilon_2} \stackrel{\triangle}{=} \{(\theta,\lambda) |\  \|\theta - \theta^*\| \leq \epsilon_1, \|\lambda - \lambda^*\| \leq \epsilon_2\}
\end{equation}
for some $\epsilon_1,\epsilon_2 > 0$.
Assuming $L(\theta, \lambda)$ is determinable for every $(\theta, \lambda)$ tuple,
the gradient search algorithm will be used as to identify a local $(\theta^*,\lambda^*)$
pair\cite{andrychowicz2016learning}:
\begin{eqnarray}
    \theta_{n+1} &=& \theta_n - \eta_1(n)\nabla_{\theta_n} (-L(\theta_n,\lambda_n)) \\
    \label{GDA1}
    &=& \theta_n + \eta_1(n)[\nabla_{\theta_n}J^R(\pi_\theta) -\lambda_n \nabla_{\theta_n}J^C(\pi_\theta)]\\ 
    \lambda_{n+1} &=& [\lambda_n + \eta_2(n)\nabla_{\lambda_n} (-L(\theta_n,\lambda_n))]_+\\
    \label{GDA2}
    &=& [\lambda_n - \eta_2(n)(J^C(\pi_\theta)-d) ]_+.
\end{eqnarray}

Here, $[x]_+$ represents $\max(0,x)$, ensuring the Lagrange multiplier remains non-negative post-update. In \eqref{GDA1}-\eqref{GDA2}, $\eta_1(n), \eta_2(n) > 0$ $\forall n$ are the predefined step-size schedules, adhering to standard step-size conditions. 
For $i=1,2$, the conditions ${\displaystyle \sum_k \eta_i(n) = \infty, \sum_k \eta_i^2(n) < \infty}$ are satisfied.

\begin{figure*}
    \centering
    \includegraphics[width=0.9\linewidth]{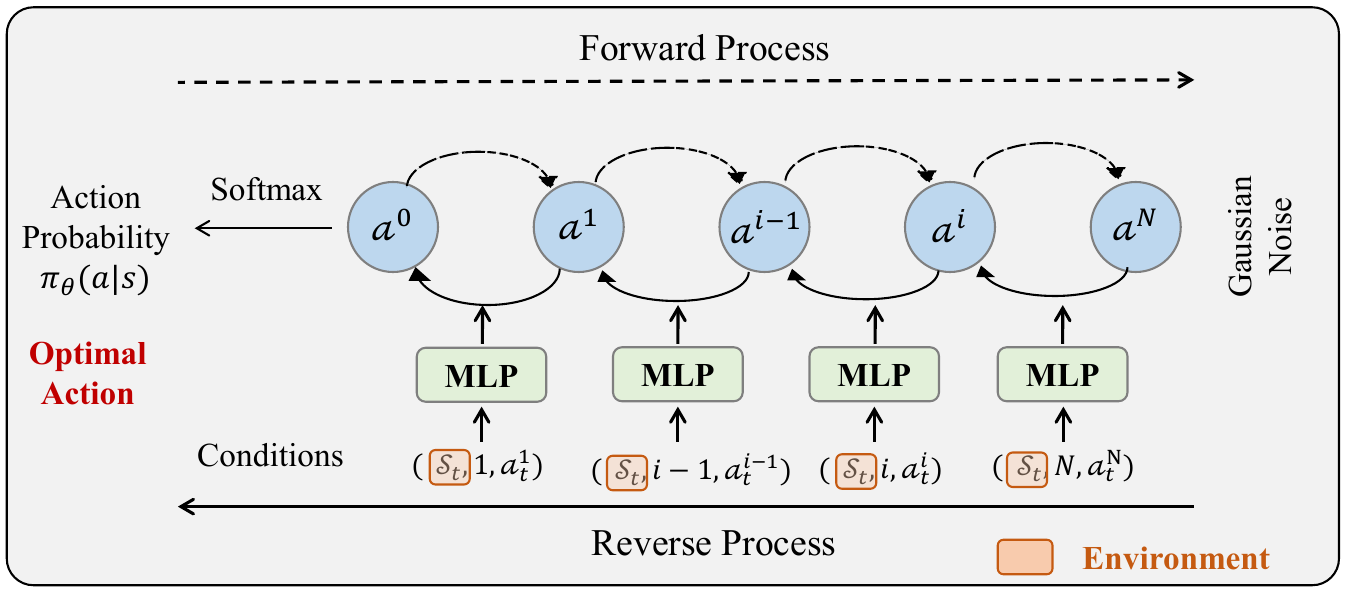}
    \caption{The conditioned diffusion process for our method.}
    \label{fig:diffusion_process}
\end{figure*}

\subsection{Problem Formulation}

The decision-making conundrum is conceptualized within the framework of a Constrained Markov Decision Process (CMDP). Commencing from an initial state \(s_0\), the AV iterates through transitions from one state \(s_t \in \mathcal{S}\) to a subsequent state \(s_{t+1} \in \mathcal{S}\) at each timestep \(t = 0, 1, \ldots, T\). This progression involves the execution of an action \(a_t \in \mathcal{A}\), consequent to which a reward \(r_t \in \mathcal{R}\) is accrued within the environmental context.

\subsubsection{State Space}
In our study, the state input $S$ of the AV comprises four main components. The first component is the ego vehicle's own state, including its position \( [x_{\text{ego}}, y_{\text{ego}}] \), velocity \( [v_{x_{\text{ego}}}, v_{y_{\text{ego}}}] \), steering angle [heading], and distance from the road boundary [$dis_{bound}$]. The second component is navigation information, for which we calculate a route from the origin to the destination and generate a set of checkpoints along the route at predetermined intervals, providing the relative distance and direction to the next checkpoint as navigation data. The third component consists of a 240-dimensional vector, characterizing the vehicle's surrounding environment in a manner akin to LiDAR point clouds. The LiDAR sensor scans the environment in a 360-degree horizontal field of view using 240 lasers, with a maximum detection radius of 50 meters and a horizontal resolution of 1.5 degrees. The final component includes the state of surrounding vehicles, such as their position and velocity information, acquired through V2X communication.

\subsubsection{Action Space}
We utilize two normalized actions to control the lateral and longitudinal motion of the target vehicle, denoted as \( \mathbf{A} = [a_1, a_2]^T \in (0,1) \). These normalized actions are subsequently translated into low-level continuous control commands: steering \( u_s \), acceleration \( u_a \), and brake signal \( u_b \) as follows:
\begin{equation}
    \label{action}
    \begin{aligned}
        &u_s = S_{\text{max}} \cdot a_1\\
        &u_a = F_{\text{max}} \cdot \max\{0, a_2\}\\ 
        &u_b = -B_{\text{max}} \cdot \min\{0, a_2\}
    \end{aligned}
\end{equation}

\noindent where \( S_{\text{max}} \) represents the maximum steering angle, \( F_{\text{max}} \) the maximum engine force, and \( B_{\text{max}} \) the maximum braking force.

\subsubsection{Reward Function}
The reward function plays a crucial role in optimizing the agent's performance.
In our study, we define the reward function as:
\begin{equation}
\label{eq:reward_function}
    R = \omega_1 r_{dis} + \omega_2 r_{v} + \omega_3 r_s
\end{equation}
The components include $r_{dis}$ for the reward based on distance covered, $r_{v}$ for the speed reward, and $r_{s}$ for the terminal reward.
% and $r_c$ as a penalty for any dangerous actions taken by the agent, which is equivalent to the cost penalty function Eq.\ref{eq:cost_function}.

\section{Methodology}
\label{section:4}
In this section, we provide a comprehensive exposition of the design of our DDM-Lag model. Initially, we develop a diffusion-based optimizer designed to generate solutions for continuous vehicle control decisions. Subsequently, we introduce a hybrid policy update method for diffusion models, integrating behavior cloning with Q-learning, and devise an Actor-Critic architecture to guide the update of the diffusion model. Finally, to augment the model's safety performance, we incorporate a safety loss as an optimization constraint, employing the Lagrangian relaxation method to address the constrained optimization problem.

\subsection{Diffusion Policy}
At any given moment, an AV makes decisions \(a\) based on the current input environmental state \(s\), with the AV's policy denoted as \(\pi_\theta(a|s)\). We model this policy using the reverse process of a conditional diffusion model.
According to the diffusion model, an optimal decision solution under the current environment can progressively increase in noise until it conforms to a Gaussian distribution, a process termed as the forward process of probability noising. Subsequently, during the reverse process of probability inference, the optimal decision generation network, denoted as \(\pi_{\boldsymbol{\theta}}(\cdot)\), functions as a denoiser that initiates with Gaussian noise and reconstructs the optimal decision solution, denoted as \(\boldsymbol{a}^0\), based on the environmental condition, \(s\). An illustration of this diffusion process is depicted in Fig.\ref{fig:diffusion_process}. 
In the following subsections, we first elucidate the forward process and then employ the reverse process of Diffusion to model the policy.

\textbf{Notation:}This paper differentiates between two categories of timesteps: diffusion timesteps, indicated by superscripts \(i \in \{1, \dots, N\}\), and trajectory timesteps, denoted by subscripts \(t \in \{1, \dots, T\}\). Within this framework, the terms autonomous vehicles and agents are used interchangeably without distinction.

\begin{figure*}
    \centering
    \includegraphics[width=0.9\linewidth]{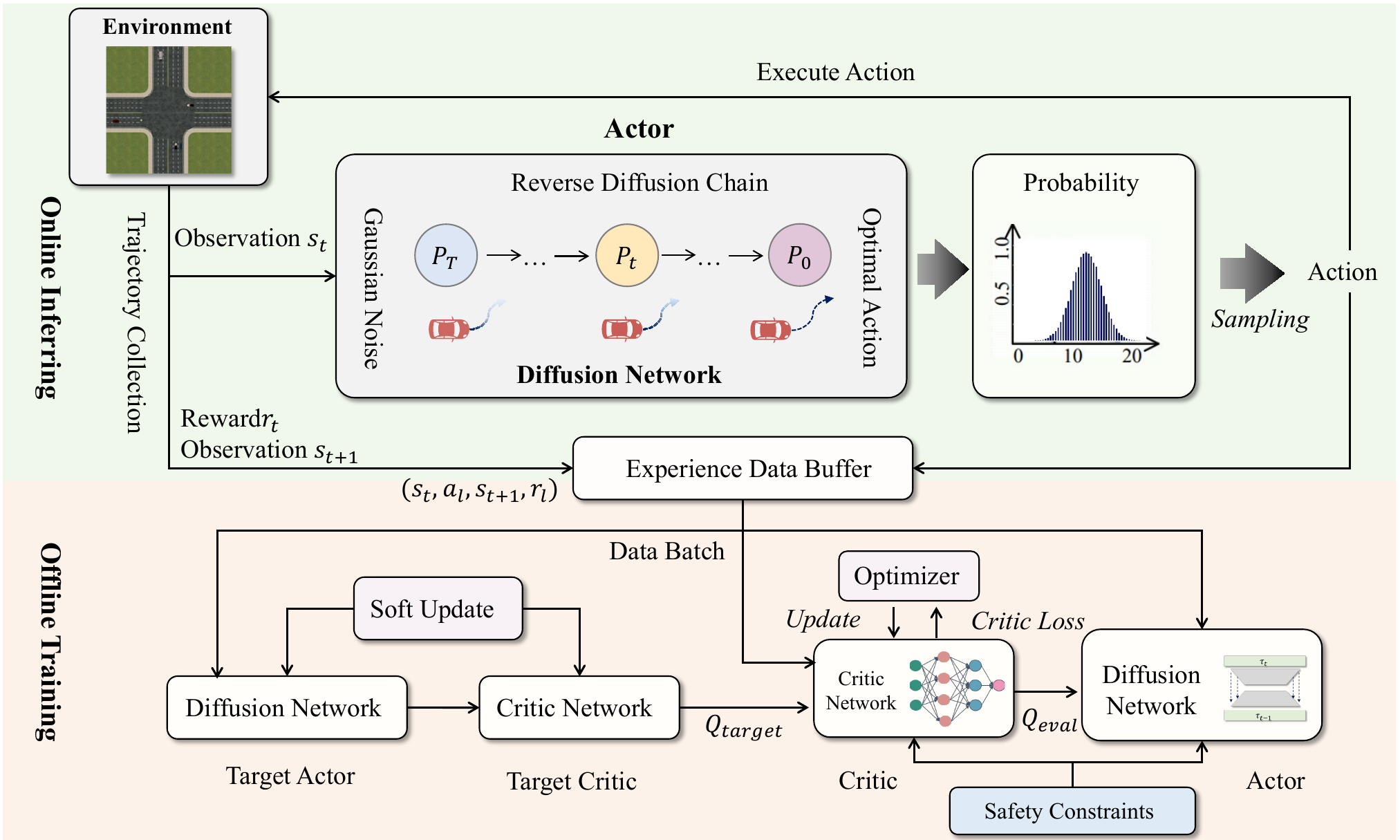}
    \caption{The framework of DDM-Lag algorithm.}
    \label{fig:algorithm_framework}
\end{figure*}
\subsubsection{The Forward Process of Probability Noising}

The decision output $\boldsymbol{a}^0 = \pi_{\boldsymbol{\theta}}(a|s) \sim \mathbb{R}^{|\mathcal{A}|}$ represents the likelihood of selecting each decision under the environmental state $s$. We denote the distribution's vector at step $i$ in the forward process as $\boldsymbol{a}^i$, maintaining the same dimensionality as $\boldsymbol{a}^0$.
To evolve the initial probability distribution $\boldsymbol{a}^0$ towards increased uncertainty, we sequentially introduce Gaussian noise at each step, yielding $\boldsymbol{a}^1, \boldsymbol{a}^2, \ldots, \boldsymbol{a}^N$. The progression from $\boldsymbol{a}^{i-1}$ to $\boldsymbol{a}^i$ follows a Gaussian distribution with a mean of $\sqrt{1-\beta_i}\boldsymbol{a}^{i-1}$ and a variance of $\beta_i\mathbf{I}$, as described by:
\begin{equation}
    q(\boldsymbol{a}^i|\boldsymbol{a}^{i-1}) = \mathcal{N}(\boldsymbol{a}^i; \sqrt{1-\beta_i}\boldsymbol{a}^{i-1}, \beta_i\mathbf{I}),
\end{equation}
where $i = 1, \ldots, N$, and $\beta_i$ is calculated based on a variance scheduling method, controlling the noise level throughout the forward process: 
\begin{equation}
    \beta _i=1-e^{-\frac{\beta _{\min}}{N}-\frac{2i-1}{2N^2}\left( \beta _{\max}-\beta _{\min} \right)}
\end{equation}

Given that $\boldsymbol{a}^i$ is solely dependent on $\boldsymbol{a}^{i-1}$, the forward process exhibits Markovian properties. The distribution of $\boldsymbol{a}^{N}$, starting from $\boldsymbol{a}^0$, is the cumulative result of all transitions $q(\boldsymbol{a}^i|\boldsymbol{a}^{i-1})$:
\begin{equation}
    q(\boldsymbol{a}^{N}|\boldsymbol{a}^0) = \prod_{t=1}^T q(\boldsymbol{a}^i|\boldsymbol{a}^{i-1}).
\end{equation}

Although not implemented practically, the forward process provides a theoretical basis to understand the connection between $\boldsymbol{a}^0$ and any $\boldsymbol{a}^i$:
\begin{equation}
    \boldsymbol{a}^i = \sqrt{\bar{\alpha}_i}\boldsymbol{a}^0 + \sqrt{1-\bar{\alpha}_i}\boldsymbol{\epsilon},
\end{equation}
where $\alpha_i = 1-\beta_i$, $\bar{\alpha}_i = \prod_{k=1}^t \alpha_k$ signifies the cumulative product up to step $t$, and $\boldsymbol{\epsilon} \sim \mathcal{N}(\mathbf{0}, \mathbf{I})$ represents standard Gaussian noise. As $t$ advances, $\boldsymbol{a}^N$ becomes entirely noise, adhering to the distribution $\mathcal{N}(\mathbf{0},\mathbf{I})$.

\subsubsection{The Reverse Process of Probability Inference}

The reverse process, also called the sampling process, aims to infer the target $\boldsymbol{a}^0$ from a noise sample $\boldsymbol{a}^N\sim\mathcal{N}(\mathbf{0},\mathbf{I})$ by removing noise from it. In our method, the purpose is to infer the optimal decision action from the noise sample.
% The transition from $\boldsymbol{a}^i$ to $\boldsymbol{a}^{t-1}$ is defined as $p_{\theta}\left( \boldsymbol{a}^{t-1}|\boldsymbol{a}^t \right)$, which cannot be calculated directly. 
% At any given moment, an autonomous vehicle must make decisions a based on its current state s, expressed as $\pi_{\theta}(a \,|\, s)$. 
We model the diffusion policy using the reverse process of conditional diffusion models:

\begin{equation}
    \textstyle\pi_{\theta}(a \,|\, s) = p_{\theta}(a^{0:N} \,|\, s) = \gN (a^N; \mathbf{0}, \mI) \prod_{i=1}^N p_{\theta}(a^{i-1}  \,|\,  a^{i}, s)
\end{equation}
where $a^0$, the end sample of the reverse chain, represents the action executed and evaluated. 
The conditional distribution $p_{\theta}(a^{i-1}  \,|\,  a^{i}, s)$ can be modeled as a Gaussian distribution $\gN(a^{i-1}; \mu_{\theta}(a^{i}, s, i), \mSigma_{\theta}(a^{i}, s, i))$. Following ~\cite{ho2020denoising}, we parameterize $p_{\theta}(a^{i-1}  \,|\,  a^{i}, s)$ as a noise prediction model, fixing the covariance matrix as $\mSigma_{\theta}(a^{i}, s, i) = \beta_i \mI$, and constructing the mean as: 
\begin{equation}
\label{eq:mean_cal}
    \mu_{\theta}(a^{i}, s, i) = \frac{1}{\sqrt{\alpha_i}} \big( a^{i} - \frac{\beta_i}{\sqrt{1 - \bar{\alpha}_i}} \rvepsilon_\theta(a^{i}, s, i) \big)
\end{equation}

The reverse diffusion chain, parameterized by $\theta$, is sampled as: 
\begin{equation} 
\label{eq:reverse_sampling}
\begin{aligned}
    \textstyle a^{i-1} \given a^i = \frac{a^i}{\sqrt{\alpha_i}} - \frac{\beta_i}{\sqrt{\alpha_i(1 - \bar{\alpha}_i)}} \rvepsilon_\theta(a^{i}, s, i) + \sqrt{\beta_i} \rvepsilon,\\
    ~~\rvepsilon \sim \mathcal{N}(\mathbf{0}, \mI),~~\text{for }i=N,\ldots,1.
\end{aligned}
\end{equation}
When $i=1$, $\rvepsilon$ is set to 0 to enhance sampling quality.
We adopt the simplified objective proposed by ~\cite{ho2020denoising} to train our conditional $\rvepsilon$-model via: 
\begin{equation}
\label{eq:bc_loss}
\begin{aligned}
    \mathcal{L}_d(\theta) = \E_{i \sim \mathcal{U}, \rvepsilon \sim \mathcal{N}(\mathbf{0}, \mI), (s, a) \sim \mathcal{D}} \\
    \sbr{|| \rvepsilon - \rvepsilon_\theta(\sqrt{\bar{\alpha}_i} a + \sqrt{1 - \bar{\alpha}_i}\rvepsilon, s, i) ||^2}
\end{aligned}
\end{equation}
where $\mathcal{U}$ is a uniform distribution over the discrete set $\{1, \dots, N\}$ and $\mathcal{D}$ denotes the offline dataset. 

\subsection{Reinforcement-Guided Diffusion Policy Learning}

The policy regularization loss in \eqref{eq:bc_loss}, $\mathcal{L}_d(\theta)$, that we employ is a behavior cloning term capable of effectively learning the behavioral patterns from expert data. However, this still falls short of bridging the gap to the complex environments and decision-making behaviors encountered in the real world, making it challenging for the model to learn strategies that surpass those in the training data. To improve the policy, we introduce guidance from the Q-value function into the backward diffusion chain during the training phase, facilitating the learning of actions with higher values through priority sampling. For this purpose, we have designed an Actor-Critic framework to guide the diffusion model in parameter updates. 
As depicted in Fig.\ref{fig:algorithm_framework}, this framework utilizes two neural networks: the policy network,  $\pi_\theta$, for decision-making, and the Q network, $Q_\phi$, for policy evaluation. The diffusion model is defined as the actor, with its policy network denoted as $\pi_\theta$, and we employ a Multilayer Perceptron (MLP) as the Critic.

During training, the Actor-Critic framework alternates between policy evaluation and policy improvement in each iteration, dynamically updating the Critic $Q_\phi$ and the Actor $\pi_\theta$. In the policy evaluation phase, we update the estimated Q-function by minimizing the L2 norm of the entropy-regularized Temporal Difference (TD) error: 
\begin{equation}
\label{equation:critic-objective-TD_error}
\begin{aligned}
    y(r_t, s_{t+1}) = r_t + \gamma \mathbb{E}_{a_{t+1}\sim \pi_\theta(\cdot|s_{t+1})} \ 
[Q_{\phi'} (s_{t+1}, a_{t+1})]
\\ -\alpha \log \pi_\theta(a_{t+1} | s_{t+1})]
\end{aligned}
\end{equation}
\begin{equation}
\label{equation:critic-objective}
L_{Q}(\phi) =\cfrac 12 \mathbb{E}_{(s_t, a_t, r_t, s_{t+1})\sim \mathcal D} [y(r_t, s_{t+1}) - Q_{{\phi}}(s_t, a_t)]^2.
\end{equation}
where $\phi'$ is the parameter of the target network $Q_{\phi'}$, and \( \alpha \) is the temperature parameter.

In the policy improvement phase, for the diffusion-based actor \( \pi_\theta \), our update target function comprises two parts: the policy regularization loss term \( \mathcal{L}_d(\theta) \) and the policy improvement objective term \( \mathcal{L}_q(\theta) \). The policy regularization loss term \( \mathcal{L}_d(\theta) \), equivalent to behavior cloning loss, is utilized for learning expert prior knowledge from human expert demonstration data. However, as it is challenging to surpass expert performance solely with this, we introduce a Q-function-based policy improvement objective term \( \mathcal{L}_q(\theta) \) during training, guiding the diffusion model to prioritize sampling high-value actions. Consequently, our policy learning objective function is expressed as: 
\begin{equation} 
\label{eq:policy_objective} 
\begin{aligned}
    \pi = \argmin_{\pi_\theta} \mathcal{L}(\theta) = \mathcal{L}_d(\theta) + \mathcal{L}_q(\theta) \\
    = \mathcal{L}_d(\theta) - \alpha \cdot \mathbb{E}_{s \sim \mathcal{D}, a^0 \sim \pi_\theta} \left[Q_{\phi}(s, a^0)\right]. 
\end{aligned}
\end{equation} 
\( a^0 \) is reparameterized by \eqref{eq:reverse_sampling}, allowing the Q-value function gradient with respect to the action to propagate backward through the entire diffusion chain.

\begin{algorithm}
\SetAlFnt{\small}
\SetKwInOut{Parameter}{Inputs}
\SetKwInOut{Output}{Outputs}
\SetKwRepeat{Repeat}{repeat}{until}
\caption{
Reinforcement Guided Diffusion Decision Model with Safe Enhancement
}
\label{alg:main}
\LinesNumbered 
\SetAlgoLined
\Parameter{
% Transition mini-batch $\mathcal{B}$,  
Expert Dataset $\mathcal{D}$}
\Output{Updated policy and critics: $\theta'$,$\phi'$,$\psi'$}
\vspace{0.2em}
\hrule
\vspace{0.2em}
Initialize policy network $\pi_{\theta}$; critic networks $Q_{\phi}$;cost critic networks $Q^C_{\psi}$; and their target networks $\pi_{\theta'}$, $Q_{\phi'}$, $Q^C_{\psi'}$; Lagrange parameter $\lambda_0 \ge 0$; Integral 
 $I\gets0$;\\
Choose tuning parameters: $K_P, K_I, K_D \geq 0$; Previous Cost: $J_{C,prev}\gets 0$;\\
\For{$Epoch = 1$ to $M$}
{
    Sample transition mini-batch $\mathcal{B} = \{(s_t, a_t, r_t, s_{t+1})\} \sim \mathcal{D}$;\\
    Initialize a random normal distribution $\boldsymbol{a}^N\sim\mathcal{N}(\boldsymbol{0},\mathbf{I})$;\\
    \For{the denoising step $t=T$ to 1}{
        Infer and scale a denoising distribution using a deep neural network;\\
        Calculate the mean $\boldsymbol{\mu}_{\boldsymbol{\theta}}$ of the reverse transition distribution $p_{\boldsymbol{\theta}}\left( \boldsymbol{a}^{i-1}|\boldsymbol{a}_i \right)$ by \eqref{eq:mean_cal};\\
        Calculate the distribution $\boldsymbol{a}^{i-1}$ by \eqref{eq:reverse_sampling} ;\\
    }
    Sample the action $\boldsymbol{a}^0_t$;\\

    \textit{\bfseries  Update lagrange multiplier $\lambda$:}\\
    Receive cost $J_C = \mathbb{E}_{\tau} [\sum_{t=0} \hat{c}_t]$;\\
    $\Delta \gets J_C-d$;\\
    $\partial \gets (J_C-J_{C,prev})_+$;\\
    $I \gets (I + \Delta)_+$;\\
    $\lambda \gets (K_P \Delta + K_I I + K_D \partial)_+$;\\
    $J_{C,prev}\gets J_C$;\\

    \textit{\bfseries  Policy learning:}\\
    Update diffusion policy parameter $\theta$:\\
    $\theta_{n+1} = \theta_n + \eta_1(n) \big[\nabla_{\theta_n}  [\mathcal{L}_d(\theta) + \mathcal{L}_q(\theta) ] -\lambda_n \nabla_{\theta_n} \mathcal{L}_q^{\lambda}(\theta) \big]$;\\
    Update $Q_{\phi}$ and $Q_{\psi}$ by \eqref{equation:critic-objective};\\
    Update target networks:\\
    $\theta' \gets \rho \theta' + (1 - \rho) \theta$, \
    $\phi' \gets \rho \phi' + (1 - \rho) \phi$,\\
    $\psi' \gets \rho \psi' + (1 - \rho) \psi$,\\
     
}
\end{algorithm}

\begin{figure*}[!htbp]
    \centering
    \includegraphics[width=0.7\linewidth]{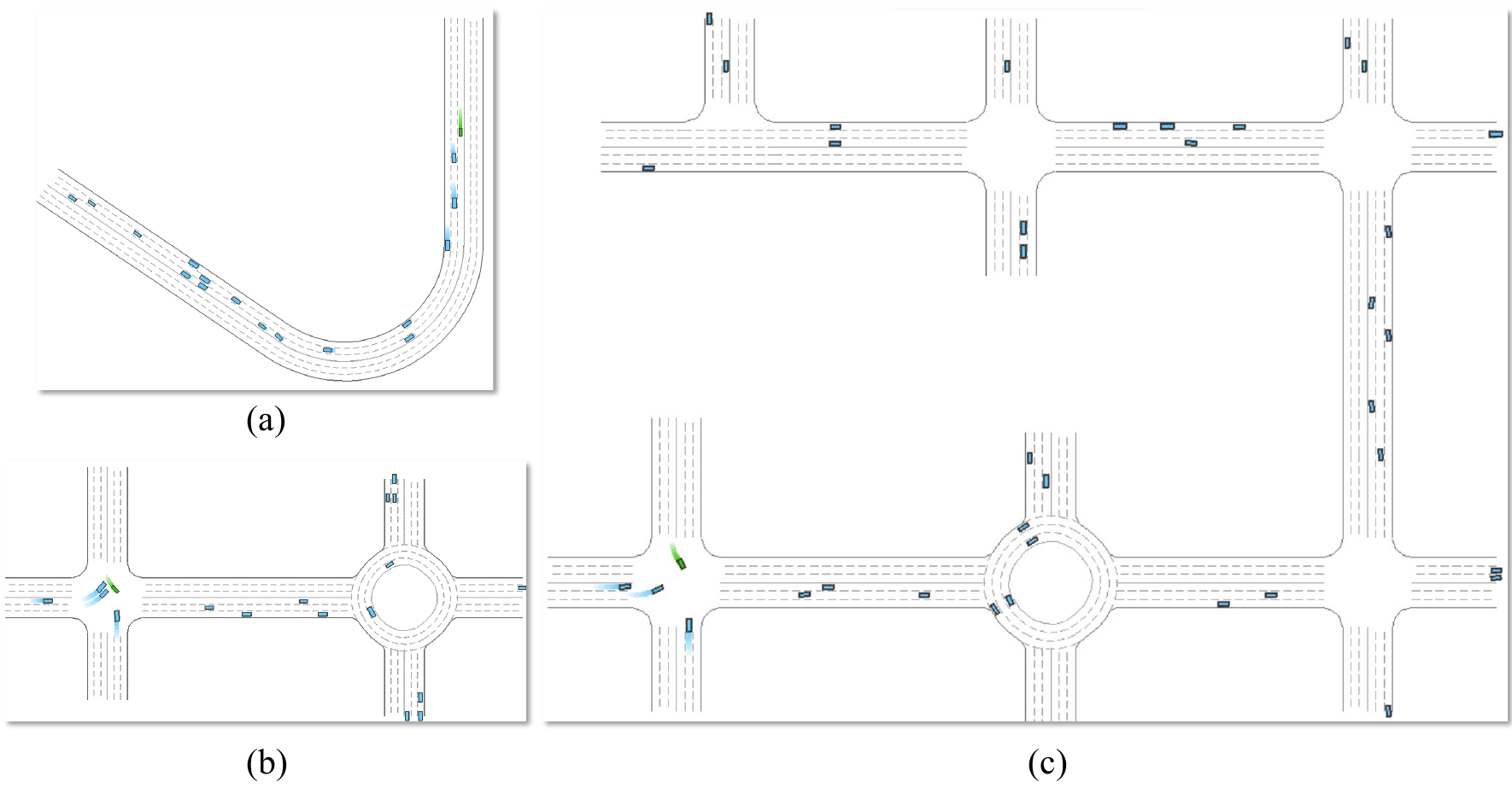}
    \caption{The three kinds of mixed scenario used for traning and testing our method, (a) scenario 1, (b) scenario 2, (c) mixed long-distance scenario 3.}
    \label{fig:scenario_exp}
\end{figure*}
\subsection{Safety Enhancement with Constrained Optimization}
Due to the inherent randomness in the Actor's exploratory process, ensuring the safety of its actions is a significant challenge, particularly in autonomous vehicle decision-making. To enhance safety in the learning and policy updating processes, we incorporate safety constraints into the policy update process, treating the original problem as a constrained optimization issue. 
The safety cost function is defined as follows:
\begin{equation}
\label{eq:cost_function}
    C = \omega'_{1} c_{1} + \omega'_{2} c_{2} + \omega'_{3} c_{3}
\end{equation}
where $c_{1}$, $c_{2}$ and $c_{3}$ are the penalty for the condition: out of road, crashing with other vehicles, crashing with other objects, respectively.

Then we apply the Lagrangian method to this optimization process. The entire optimization problem becomes: 
\begin{equation} 
\label{eq:lagrangian} 
\theta^* = \mathop{\arg\max}_{\theta}\ \mathop{\min}_{\lambda\ge0} \mathbb{E} \{ (\sum_{t=0} \gamma^t R_t) -\lambda [(\sum_{t=0} \gamma^t \hat{C}_t)-C] \} 
 \end{equation} 
Here, \( \theta \) and \( \lambda \) are updated through policy gradient ascent and stochastic gradient descent (SGD)\cite{amari1993backpropagation}.
% as shown in Eq.\ref{GDA1}-Eq.\ref{GDA2}. 
We then introduce a safety-assessing Critic \( Q_{\psi}^C \) to estimate the cumulative safety constraint value \( \sum_{t'=t} \gamma^{(t-t')} C_t \). 
With the reward replaced by the safety constrain, the safety-critic network can be optimized by \eqref{equation:critic-objective}.
For the actor $\pi_\theta$, the safety constraint violation minimization objective can be written as: 
\begin{equation} 
\label{eq:constrain-objective}
 L_{q}^{\lambda}(\theta) = \mathbb{E}_{s_t\sim \mathcal D, a_t \sim \pi_\theta(\cdot|s_t)} [Q^C_\psi(s_t, a_t) - C]
\end{equation} 
Now, by combining the original policy improvement objective \eqref{eq:policy_objective} and the safety constraint minimization optimization objective \eqref{eq:constrain-objective}, we derive our final policy update objective: 
\begin{equation}
\label{eq:policy_objective_final} 
\pi = \argmin_{\pi_\theta} \mathcal{L}(\theta) = \mathcal{L}_d(\theta) + \mathcal{L}_q(\theta) - \lambda \mathcal{L}_q^{\lambda}(\theta)
\end{equation}

However, directly optimizing the Lagrangian dual during policy updates can lead to oscillation and overshooting, thus affecting the stability of policy updates. From a control theory perspective, the multiplier update represents an integral control. Therefore, following \cite{stooke2020responsive}, we introduce proportional and derivative control to update the Lagrangian multiplier, which can reduce oscillation and lower the degree of action violation. Specifically, we use a PID controller to update \( \lambda \), as follows:
 \begin{equation}
 \label{equation:lambda-update-rule} 
 \begin{aligned}
     \lambda \gets K_p \delta + K_i \int_{i=1}^{k} \delta di + K_d \cfrac{\delta}{di}, \\
     \delta = \mathbb{E}_{\tau} [\sum_{t=0} \hat{c}_t] - C  
 \end{aligned}
\end{equation} 
where we denote the training iteration as \( i \), and \( K_p \), \( K_i \), \( K_d \) are the hyper-parameters. Optimizing \( \lambda \) with ~\eqref{eq:lagrangian} reduces to the proportional term in ~\eqref{equation:lambda-update-rule}, while the integral and derivative terms compensate for the accumulated error and overshoot in the intervention occurrence.
The whole training procedure of DDM-Lag is summarized in Algorithm \ref{alg:main}.

\section{Experiment and Evaluation}
\label{section:5}
In this section, the detailed information of the simulation environment and our models will be introduced. Sequently, the experiments results are analyzed.

\subsection{Experiments and Baselines}
\textbf{Environments.}
We conducted our experiments, including data collection, model training, and testing, in the MetaDrive Simulator \cite{li2022metadrive}, which is based on OpenAI Gym Environment and allows for the creation of various traffic scenarios.
To comprehensively assess the performance of our method, we established three distinct composite traffic scenarios for training and testing our algorithm, as depicted in Fig.\ref{fig:scenario_exp}. The first two scenarios are designed for short-distance tests, while the third scenario is for long-distance testing. The first traffic scenario includes straight roads and a segment of curved road to evaluate the basic driving capabilities of AVs on urban roads. In the second scenario, we set up a mixed scenario comprising unsignalized intersections and roundabouts, where the autonomous vehicle is required to execute challenging driving maneuvers such as unprotected left turns. The third scenario is designed to evaluate the performance of autonomous vehicles over longer driving distances, featuring multiple complex intersections and roundabouts. Within each scenario, we manipulate the background traffic flow density to control the complexity of the scenario, specifically setting a traffic density of 0.1 as low density and 0.2 as high density.

\textbf{Dataset Collection.}
An offline expert dataset is indispensable for training the diffusion model. As illustrated in Figure \ref{fig:overall_procedure}, we utilize the Soft Actor-Critic (SAC) with the Lagrangian algorithm\cite{ha2020learning} and human expertise to collect experience data. This expert experience data encompasses trajectories and rewards from diverse environments, which are stored in the data buffer and constitute the offline dataset. For every scenario, differing in difficulty and traffic density, 10,000 trajectories are respectively collected with the expert agent.

\textbf{Baselines.}
We selected several competitive baseline algorithms that have been widely employed in recent years for safe reinforcement learning and autonomous driving decision control. These include the classical Behavior Cloning algorithm (BC), TD3+BC \cite{fujimoto2021minimalist}, Decision Transformer (DT) \cite{chen2021decision}, IQL \cite{kostrikov2021offline}, Deep Q-learning with a diffusion policy (Diffusion Q-learning), and Q learning with CAVE policy network (CAVE-QL). These methods encompass offline RL, online RL, sequence modeling-based RL, and decision modeling methods based on other generative approaches, providing a comprehensive benchmark for evaluating the performance of our method.

\begin{table}[!htbp]
    \centering
    \caption{The hyperparameter setting of our work}
    \label{tab:training_hyperparameter}
    \begin{tabular}{c c c c}
        \toprule
        Symbol & Definition & Value \\
        \midrule
        $B_s$ & Batch Size & 256  \\
        $l_a$ & actor learning rate & 0.001  \\
        $l_c$ & critic learning rate & 0.99 \\
        $\lambda$ & Weight of the Lagrangian term & 0.75\\
        $\gamma$ & Discount factor for the reward  & 0.99\\
        $\tau$ &  Soft update coefficient of the target network & 0.005 \\
        \bottomrule
    \end{tabular}
\end{table}

\subsection{Implementation Details}
For each model,the training timesteps is $20K$ , batch size is 512 and the cost uplimit $C$ for the cost function  \eqref{eq:cost_function}  is set as 10.  Our diffusion model is built based on a 3-layer MLP with 256 hidden units for all networks. As for BCQ and Diffusion Q-learning baseline, the policy network is set as same as our diffusion actor. For the diffusion model, the number of diffusion steps $N$ is set as 5. The $K_p$, $K_i$ and $K_d$ for the PID controller are set as 0.1, 0.003, 0.001, respectively.
We set the following values for the reward function: $r_{dis}=1$, $r_{v}=0.1$, $r_{s}=10$, $r_{c1}=5.0$, $r_{c2}=5.0$ and $r_{c3}=5.0$. The coefficient of each reward term is set as 1.
In the safety cost function, $c_{1}=1.0$, $c_{2}=5.0$ and $c_{3}=5.0$. The coefficient of each cost function term is set as 1. 
Other parameters are shown in Tab.\ref{tab:training_hyperparameter}.
All experiments are conducted in a computation platform with Intel Xeon Silver 4214R CPU and  NVIDIA GeForce RTX 3090 GPU.

\subsection{Results Analysis}
\begin{table*} 
\centering
\caption{The Average Safety Cost of Different Algorithms in Testing.}
\label{tab:safety_cost}
\begin{tabular}{@{}l||cccccc|c@{}}
\toprule
\textbf{Task} & \textbf{BC} & \textbf{Diffusion QL} & \textbf{CVAE-QL} & \textbf{TD3+BC} & \textbf{DT} & \textbf{IQL} & \textbf{DDM-Lag} \\
\hline
Sce.1-Den.1 & 13.654 & 8.919 & 18.043 & 8.564 & 12.610 & 3.865 & \textbf{0.720} \\
Sce.1-Den.2 & 30.283 & 31.146 & 17.293 & 34.604 & 8.953 & 17.958 & \textbf{0.874} \\
Sce.2-Den.1 & 18.625 & 22.825 & 24.121 & 30.713 & 14.147 & 17.536 & \textbf{0.789} \\
Sce.2-Den.2 & 22.712 & 30.096 & 31.054 & 33.135 & 23.789 & 24.100 & \textbf{0.980} \\
Sce.3-Den.1 & 57.955 & 75.110 & 42.952 & 81.654 & 26.546 & 50.041 & \textbf{2.103} \\
Sce.3-Den.2 & 60.049 & 86.197 & 80.586 & 57.479 & 55.393 & 64.472 & \textbf{1.954} \\
\hline
\textbf{Average} & 33.880 & 42.382 & 35.675 & 41.025 & 23.573 & 29.662 & \textbf{1.237} \\
\bottomrule
\end{tabular}
\end{table*}

\begin{table*} 
\centering
\caption{The Mean Reward of Different Algorithms in Testing.}
\label{tab:mean_reward}
\begin{tabular}{@{}l||cccccc|c@{}}
\toprule
\textbf{Task} & \textbf{BC} & \textbf{Diffusion QL} & \textbf{CVAE-QL} & \textbf{TD3+BC} & \textbf{DT} & \textbf{IQL} & \textbf{DDM-Lag} \\ 
\hline
Sce.1-Den.1 & 105.6 & 132.8 & 188.6 & 210.5 & 184.4 & 209.5 & \textbf{230.6} \\
Sce.1-Den.2 & 107.9 & 47.3 & 114.3 & 120.8 & 137.0 & 140.0 & \textbf{205.7} \\
Sce.2-Den.1 & 118.5 & 127.3 & 173.1 & 197.7 & 175.9 & 183.3 & \textbf{206.7} \\
Sce.2-Den.2 & 92.3 & 92.4 & 142.0 & 149.2 & 126.5 & 148.9 & \textbf{171.5} \\
Sce.3-Den.1 & 277.7 & 260.9 & 467.5 & 281.8 & 392.7 & 446.5 & \textbf{491.3} \\
Sce.3-Den.2 & 240.4 & 275.0 & 269.8 & 370.3 & 322.7 & 329.8 & \textbf{456.6} \\
\hline
\textbf{Average} & 157.1 & 156.0 & 225.9 & 221.7 & 223.2 & 243.0 & \textbf{293.7} \\
\bottomrule
\end{tabular}
\end{table*}

\begin{table*} 
\centering
\caption{The Average Safe Running Length of Different Algorithms in Testing.}
\label{tab:running_length}
\begin{tabular}{@{}l||cccccc|c@{}}
\toprule
\textbf{Task} & \textbf{BC} & \textbf{DiffusionQL} & \textbf{CVAE-QL} & \textbf{TD3+BC} & \textbf{DT} & \textbf{IQL} & \textbf{DDM-Lag} \\ 
\hline
Sce.1-Den.1 & 313.7 & 370.8 & 546.8 & 601.0 & 515.6 & 588.0 & \textbf{643.6} \\
Sce.1-Den.2 & 288.7 & 153.7 & 322.4 & 325.8 & 371.4 & 360.2 & \textbf{522.5} \\
Sce.2-Den.1 & 413.0 & 437.0 & 640.8 & 687.2 & 602.6 & 624.5 & \textbf{712.3} \\
Sce.2-Den.2 & 362.7 & 401.5 & 579.9 & 602.6 & 508.6 & 600.2 & \textbf{668.9} \\
Sce.3-Den.1 & 1093.0 & 1190.2 & 1833.5 & 1164.7 & 1575.4 & 1793.9 & \textbf{1915.8} \\
Sce.3-Den.2 & 955.3 & 1107.2 & 1142.6 & 1504.6 & 1215.2 & 1383.1 & \textbf{1780.5} \\
\hline
\textbf{Average} & 571.0 & 610.1 & 844.3 & 814.3 & 798.1 & 891.7 & \textbf{1040.6} \\
\bottomrule
\end{tabular}
\end{table*}

% \begin{table} 
% \centering
% \caption{Interaction Performance Evaluation Results of Different Algorithms.}
% \label{tab:interaction_performance_metrics}
% \begin{tabular}{@{}l||ccc|c@{}}
% \toprule
%  & \textbf{BC} & \textbf{TD3+BC} & \textbf{Diffusion QL} & \textbf{DDM-Lag} \\ 
% \hline
% \textbf{TTC-Sce.1} & 2.71 & 1.82 & 0.48 & 2.4 \\
% \textbf{Ave Speed(m/s)-Sce.1} & 6.32 & 13.19 & 22.98 & 12.26 \\
% \textbf{PET-Sce.2} & 5.04 & 7.74 & 4.29 & 3.34 \\
% \textbf{Ave Speed(m/s)-Sce.2} & 5.22 & 27.28 & 9.05 & 8.2 \\
% \bottomrule
% \end{tabular}
% \end{table}

\subsubsection{Basic Performance Analysis}
Our method was evaluated alongside several baseline algorithms across a variety of scenarios with different traffic densities. Our evaluation metrics include Mean Reward, Safety Cost, and Safe Running Length. Mean Reward is calculated using \eqref{eq:reward_function} and assesses the average performance of different algorithms across multiple evaluation iterations; Safety Cost, determined by \eqref{eq:cost_function}, quantifies the incidence of safety violations during autonomous driving; Safe Running Length evaluates the average duration an autonomous vehicle can drive safely across different distances. 

For each test, if the vehicle exhibits any hazardous driving behavior, such as collisions, lane departures, or leaving the lane, the record is reset to zero. Overall, an algorithm that achieves a higher Mean Reward, lower Safety Cost, and longer Safe Running Length demonstrates superior comprehensive performance.
Animations demonstrating  cases from DDM-Lag  can be accessed at the site.\footnote{See \url{https://drive.google.com/drive/folders/1SypbnDVqn4xD85s-UjRkxkcI0Rtd2OXs}}

\textbf{Safety Analysis.}
In autonomous driving, safety is paramount. The safety cost for different algorithms under various scenarios is presented in Tab.\ref{tab:safety_cost}, where a lower safety cost signifies fewer risky decisions and higher algorithmic safety. It is observed that Diffusion QL and TD3+BC exhibit the highest average safety cost across the three scenarios. DT and IQL algorithms perform slightly better than other baselines, while our method, owing to the explicit inclusion of safety constraints, significantly reduces safety violations compared to other baselines, achieving the lowest safety cost. Additionally, across different scenarios with varying densities, our method maintains safety cost within a controlled range, demonstrating adequate reliability and stability. Across the scenarios, it is notable that most algorithms exhibit better safety performance in low-density situations, as high-density scenarios may introduce more potential collisions and hazards.

\textbf{Comprehensive Performance Analysis.}
Given that safety cost is related to the running length of autonomous vehicles in scenarios, the comprehensive performance of different algorithms requires further analysis. We assess the overall performance of algorithms in test scenarios using average reward and Safe Running Length, with results presented in Tab.\ref{tab:mean_reward} and Tab.\ref{tab:running_length}. The tables reveal that BC and DiffusionQL have the lowest average rewards, but it is noteworthy that these two algorithms exhibit higher Average Safe Running Length compared to other baselines. This indicates that during testing, these algorithms spend more time on decision-making inference but achieve poorer performance, suggesting suboptimal decision efficiency and overall performance. The remaining four baselines have closely matched Average Reward performances, with IQL performing slightly better. In comparison, our method demonstrates the best performance across different scenarios.

\subsubsection{Interaction Ability Analysis}
\begin{figure}
    \centering
    \includegraphics[width=0.9\linewidth]{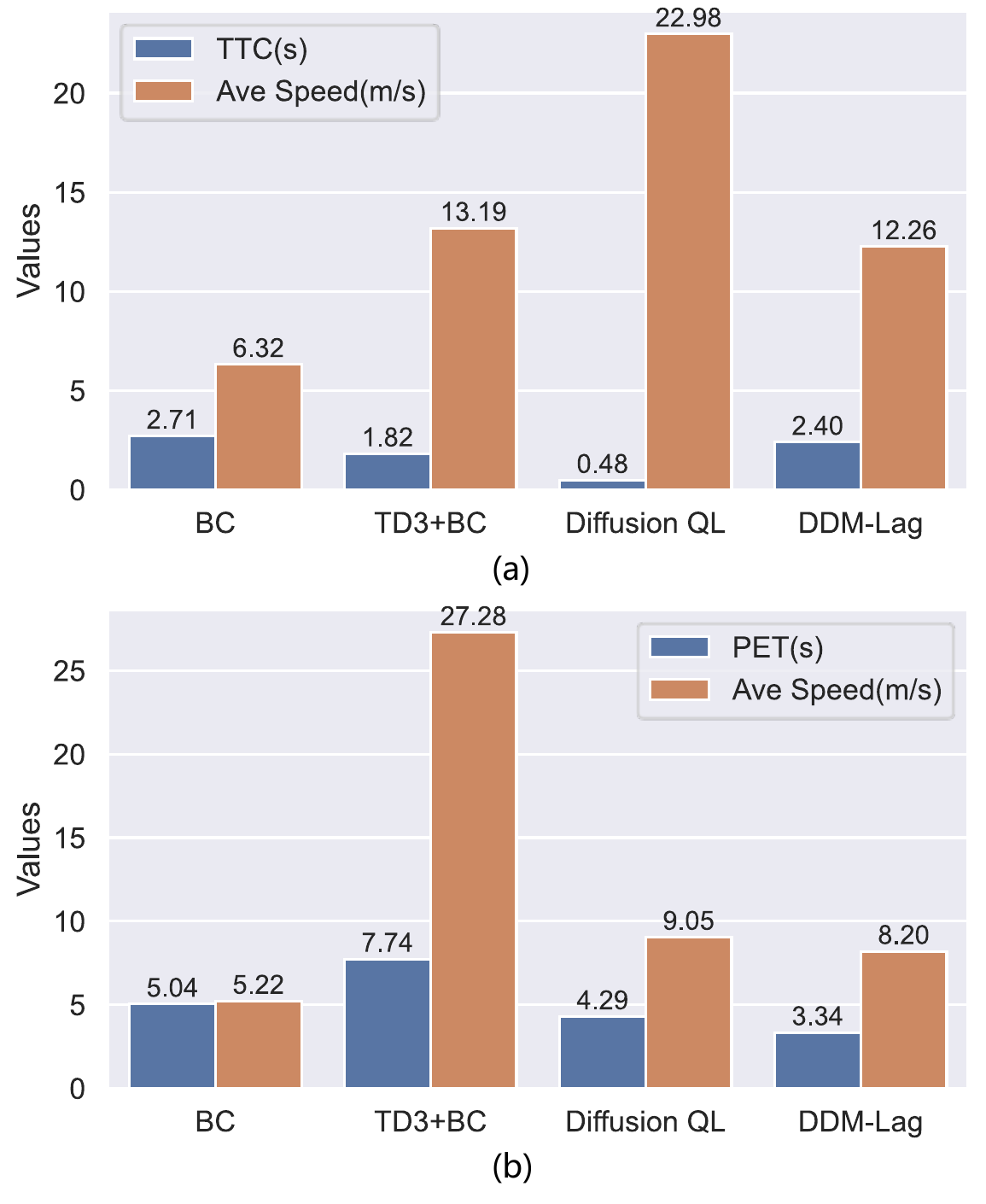}
    \caption{Interaction performance evaluation results for different scenarios, (a) scenario 1, (b) scenario 2.}
    \label{fig:interaction_performance}
\end{figure}

Beyond basic decision-making performance, the ability of algorithms to interact with human-driven vehicles (HVs) in complex scenarios warrants attention. Algorithms that perform well on evaluation metrics may not necessarily be suitable for interaction with human drivers on the road due to potentially aggressive or overly conservative behaviors that are difficult for humans to comprehend. We employ two of the most widely used indicators in traffic safety engineering and driving interaction evaluation: Time to Collision (TTC) and Post Encroachment Time (PET). Time to Collision refers to the estimated time remaining before a collision would occur between the subject vehicle and a target vehicle, assuming no change in their speeds or directions. Post-Encroachment Time (PET) is an indicator of conflict severity that measures the time difference between one vehicle leaving and another vehicle entering a common area of potential conflict, making it particularly apt for assessing vehicular interactions in intersection scenarios.

Given their applicability for analyzing micro-interaction behaviors, we focus our analysis on two short-distance scenarios, Scenario 1 and Scenario 2, with TTC applicable to Scenario 1 and PET to Scenario 2. We calculated the average minimum TTC and average PET for our method and two baseline algorithms, along with the corresponding driving speeds, as shown in Fig.\ref{fig:interaction_performance}.

In Scenario 1, the Diffusion QL algorithm exhibits the lowest TTC and highest average speed, suggesting a tendency towards more aggressive and risky driving behaviors. Conversely, the BC algorithm shows a lower average speed, indicating a possible conservative bias in the learned decisions. This demonstrates that while BC and Diffusion QL may excel in evaluation metrics, they might not be directly applicable to real-world traffic decision-making due to their extreme behavioral tendencies. In contrast, DDM-Lag manages to maintain a larger TTC while balancing throughput efficiency, indicating a certain advantage.

In Scenario 2, the TD3+BC algorithm shows the smallest PET and highest average speed, implying a higher interaction risk at intersections. Similarly, the BC algorithm exhibits the largest PET but the slowest average speed, mirroring its conservative performance in Scenario 1. Our DDM-Lag algorithm, however, performs better by maintaining a good balance between safety and efficiency.

\begin{table*}
\centering
\caption{Ablation study. We conduct an ablation study to compare our DDM-Lag model with different methods.}
\label{tab:ablation_study_reward}
\begin{tabular}{@{}l||cccc|c@{}}
\toprule
\textbf{Tasks} & \textbf{A2C} & \textbf{BC-Diffusion} & \textbf{A2C-Lag} & \textbf{A2C-Diffusion} & \textbf{DDM-Lag} \\
\hline
Sce.1-Den.1 & 166.8 & 181.7 & 228.0 & 224.4 & \textbf{230.6} \\
Sce.1-Den.2 & 148.8 & 162.1 & 203.4 & 200.2 & \textbf{205.7} \\
Sce.2-Den.1 & 201.0 & 143.5 & 192.7 & 201.1 & \textbf{206.7} \\
Sce.2-Den.2 & 166.8 & 119.1 & 165.7 & 167.0 & \textbf{171.5} \\
Sce.3-Den.1 & 217.6 & 444.3 & 477.1 & 479.2 & \textbf{491.3} \\
Sce.3-Den.2 & 202.3 & 412.9 & 443.4 & 445.4 & \textbf{456.5} \\
\hline
\textbf{Average} & 183.9 & 244.0 & 285.0 & 286.2 & \textbf{293.7} \\
\bottomrule
\end{tabular}
\end{table*}

\subsection{Ablation Study}
In this ablation study, we conducted experiments on data from the most complex scenarios to explore the impact of employing diffusion models as a policy representation and the addition of the Lagrangian safety enhancement module on the overall method. We compared four algorithms for the ablation study: Advantage Actor Critic (A2C,without the diffusion module and Lagrangian safety module), Diffusion-A2C (without the Lagrangian safety module), A2C-Lag (without the diffusion module), and Diffusion-BC. The comparsion results are shown in Tab.\ref{tab:ablation_study_reward}.

It is evident that our method, in comparison to both A2C and A2C-Lag, significantly enhances the model's stability and average reward through the incorporation of the diffusion module, resulting in superior average performance. Moreover, when comparing the A2C+diffusion and BC+Diffusion algorithms, the addition of the Lagrangian module to our method further elevates the safety performance of the model. The ablation experiments substantiate that the various components of our proposed method synergistically operate to yield commendable performance.

\section{Conclusion}
\label{section:6}
Decision-making processes are fundamental to the operational integrity and safety of autonomous vehicles (AVs). Contemporary data-driven decision-making algorithms in this domain exhibit a discernible potential for enhancements.

In this study, we introduce DDM-Lag, a  diffusion-based decision-making model for AVs, distinctively augmented with a safety optimization constraint. A key point in our approach involves the integration of safety constraints within CMDP to ensure a secure action exploration framework. Furthermore, we employ a policy optimization method based on Lagrangian relaxation to facilitate comprehensive updates of the policy learning process. 
The efficacy of the DDM-Lag model is evaluated in different driving tasks. Comparative analysis with baseline methods reveals that our model demonstrates enhanced performance, particularly in the aspects of safety and comprehensive operational effectiveness.

Looking ahead, we aim to further refine the inference efficiency of the DDM-Lag model by fine-tuning its hyperparameters. We also plan to explore and integrate additional safety enhancement methodologies to elevate the safety performance of our model. Moreover, the adaptability and robustness of our model will be subjected to further scrutiny through its application in an expanded array of scenarios and tasks. 

\bibliographystyle{IEEEtran}  
\bibliography{reference}

\end{document}